\RequirePackage{pdf14}
\documentclass[letterpaper, 10pt, conference]{ieeeconf}
\usepackage[table]{xcolor}
\usepackage[hidelinks]{hyperref}
\usepackage[a-1b]{pdfx}

\IEEEoverridecommandlockouts                              

\let\labelindent\relax
\usepackage{graphicx}
\usepackage{subcaption}
\usepackage{enumitem}
\usepackage[switch]{lineno}

\usepackage{listings}
\usepackage{booktabs}
\usepackage{cite}
\usepackage{csquotes}
\usepackage{amsmath,amssymb,amsfonts}
\usepackage{algorithmic}
\usepackage{graphicx}
\usepackage{textcomp}
\usepackage{verbatim}
\usepackage{fancybox}
\usepackage{multirow}
\usepackage{tabularx}
\usepackage[nolist]{acronym}

\usepackage[capitalize]{cleveref}
\crefname{section}{Sect.}{sections}
\Crefname{section}{Section}{Sections}

\newcommand\YAMLcolonstyle{\color{red}\mdseries}
\newcommand\YAMLkeystyle{\color{black}\mdseries}
\newcommand\YAMLvaluestyle{\color{blue}\mdseries}

\makeatletter

\newcommand\language@yaml{yaml}
\expandafter\expandafter\expandafter\lstdefinelanguage
\expandafter{\language@yaml}
{
  keywords={True,False,null,y,n},
  keywordstyle=\color{darkgray}\bfseries,
  basicstyle=\ttfamily\footnotesize,                                 
  sensitive=false,
  comment=[l]{\#},
  morecomment=[s]{/*}{*/},
  commentstyle=\color{purple}\ttfamily,
  stringstyle=\YAMLvaluestyle\ttfamily,
  moredelim=[l][\color{orange}]{\&},
  moredelim=[l][\color{magenta}]{*},
  moredelim=**[il][\YAMLcolonstyle{:}\YAMLvaluestyle]{:},   
  morestring=[b]',
  morestring=[b]",
  literate =    {---}{{\ProcessThreeDashes}}3
                {>}{{\textcolor{red}\textgreater}}1     
                {|}{{\textcolor{red}\textbar}}1 
                {\ -\ }{{\mdseries\ -\ }}3,
}
\lst@AddToHook{EveryLine}{\ifx\lst@language\language@yaml\YAMLkeystyle\fi}
\makeatother

\expandafter\expandafter\expandafter\lstdefinelanguage
\expandafter{scenic}
{
  morekeywords={class, visible, on, right, left, of, offset, by, facing, relative, to, at, ahead, with, deg}
}

\definecolor{codegreen}{rgb}{0,0.6,0}
\definecolor{codegray}{rgb}{0.5,0.5,0.5}
\definecolor{codepurple}{rgb}{0.58,0,0.82}
\definecolor{backcolour}{rgb}{0.95,0.95,0.92}

\lstdefinestyle{mystyle}{
  backgroundcolor=\color{backcolour},
  commentstyle=\color{codegreen},
  keywordstyle=\color{magenta},
  numberstyle=\tiny\color{codegray},
  stringstyle=\color{codepurple},
  basicstyle=\ttfamily\footnotesize,
  breakatwhitespace=false,
  breaklines=true,
  captionpos=b,
  keepspaces=true,
  showspaces=false,
  showstringspaces=false,
  showtabs=true,
  tabsize=2,
  framextopmargin=1pt,
  framexbottommargin=1pt,
  framexleftmargin=2pt,
  rulecolor=\color{gray},
  frame=single,
}
\newboolean{draft}
\setboolean{draft}{true}

\ifthenelse{\boolean{draft}}{
\newcommand{\clg}[1]{{\color{green}\textbf{(CLG:} #1\textbf{)}}}
\newcommand{\afs}[1]{{\color{orange}\textbf{(AA:} #1\textbf{)}}}
\newcommand{\ct}[1]{{\color{blue}\textbf{(CT:} #1\textbf{)}}}
\newcommand{\todo}[1]{{\color{red}(TODO: #1)}}
\newcommand{\missingfigure}[1]{{\color{red}MISSING FIGURE: #1}}
}
{
\newcommand{\clg}[1]{}
\newcommand{\afs}[1]{}
\newcommand{\ct}[1]{}
\newcommand{\todo}[1]{}
\newcommand{\missingfigure}[1]{}
}

\newcommand{\artifacturl}{\url{https://github.com/squaresLab/GzScenic}}

\title{\LARGE \bf
GzScenic: Automatic Scene Generation for Gazebo Simulator
}
\author{Afsoon Afzal,$^{1}$ Claire Le Goues,$^{1}$ and Christopher S. Timperley$^{1}$
\thanks{$^{1}$All authors are with School of Computer Science,
        Carnegie Mellon University, Pittsburgh, PA
        {\tt\small afsoona@cs.cmu.edu, clegoues@cs.cmu.edu, ctimperley@cmu.edu}}%
}

\begin{document}

\maketitle
\thispagestyle{empty}
\pagestyle{empty}

\begin{abstract}

Testing robotic and cyberphysical systems in simulation require
specifications of the simulated environments (i.e., scenes).
The Scenic domain-specific language 
provides a high-level probabilistic programming language that allows
users to specify scenarios for simulation. Scenic automatically 
generates concrete scenes that can be rendered by simulators.
However, Scenic is mainly designed for autonomous vehicle simulation
and does not support the most popular general-purpose simulator: Gazebo.
In this work, we present GzScenic; a tool that automatically generates
scenes for simulation in Gazebo. GzScenic automatically generates both
the models required for running Scenic on the scenarios, and the models
that Gazebo requires for running the simulation.

\end{abstract}

\section{Introduction}
\label{sec:intro}


As robots are deployed to new environments with greater levels of
autonomy, inevitable software defects may lead to unintentional and
potentially catastrophic outcomes.
It is now more important than ever to systematically test robotic
systems as extensively as possible to identify and eliminate defects
before those systems are deployed to the field.

Prior studies have suggested simulation-based testing as a promising
technique for revealing defects that is vastly cheaper, safer, and
more scalable than field testing~\cite{SotiropoulosNavigationBugs2017,TimperleyArdu2018,Robert2020,Gladisch2019,AfzalICST20}.
While simulation suffers from certain limitations and only provides
an abstraction of the physical world~\cite{AfzalICST21},
it allows systems to be systematically tested under a wide array of
environments, conditions, and scenarios that would otherwise be
difficult or expensive to replicate in the field.

A crucial aspect of simulation-based testing
is the generation of interesting, potentially fault-revealing scenarios
that expose the system to corner cases and undertested inputs.
We define a scenario as the description of a scene (i.e., the environment)
and an accompanying mission that the system under test (SUT)
should perform in the specified scene.
Manually generating such scenes and missions can be time consuming and difficult~\cite{AfzalICST21}.

\begin{figure}
\centering
    \begin{subfigure}{0.9\columnwidth}
    \small
    \begin{lstlisting}[language=scenic]
ego = Car

spot = OrientedPoint on visible curb
badAngle = Uniform(1.0, -1.0) * Range(10, 20) deg
parkedCar = Car left of (spot offset by -0.5 @ 0), facing badAngle relative to roadDirection
\end{lstlisting}
    \label{fig:scenic-scenario}
    \caption{\small A scenario description, written in the Scenic language, detailing a scene that contains a badly parked car.}
    \vspace{2mm}
    \end{subfigure}

    \begin{subfigure}{0.9\columnwidth}
    \label{fig:scenic-scene}
    \centering
	\includegraphics[width=0.9\columnwidth]{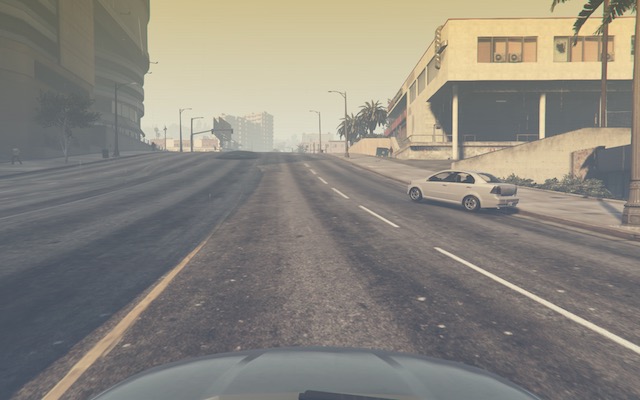}
    \caption{\small A scene that was generated by Scenic according to the scenario above using
    the GTA V engine~\cite{scenic}.}
    \end{subfigure}
    \caption{\small An exemplary Scenic scenario, and the generated
    simulation scene.}
    \label{fig:scenic}
\end{figure}

In recent years, researchers have proposed tools and domain-specific
languages (DSLs) to facilitate the construction of testing
scenarios~\cite{scenic,paracosm,kluck2018}.
One of the most prominent such DSLs is Scenic~\cite{scenic},
a language designed for creating simulation scenarios for autonomous vehicles.
Using Scenic, users can describe a scenario
of interest for the SUT, which is automatically parsed by the Scenic
tool to generate a plausible scene and mission that satisfy the user-specified
constraints of that scenario. The generated scene
and mission are then executed in the supported simulators to execute
the test. \Cref{fig:scenic} shows an
example scenario that is realized in the GTA V~\cite{gta} simulator.

Although Scenic provides a powerful language and tool that
simplifies the process of creating and running simulated test scenarios,
it only supports domain-specific simulators in the autonomous vehicle
sector, and is not compatible with Gazebo; the most popular, general-purpose
robotic simulator~\cite{gazebo}.
Gazebo is commonly used for simulation of systems developed using the
popular Robot Operating System (ROS) framework~\cite{ros}, and has been applied
to robots that span a wide variety of sectors such as unmanned aerial and ground vehicles,
agriculture robots, and industrial robots.

In this work, we introduce GzScenic; a tool that automatically
generates simulation scenes in Gazebo from a scenario provided in
Scenic's DSL.
Using GzScenic, developers can specify their desired
testing scenarios in Scenic's DSL without the need to manually
pre-define their models in Scenic, and automatically generate
complex scenes that satisfy the constraints of their scenario.
GzScenic automatically transfers the generated scenes to Gazebo
without the need for manual translation.
Furthermore, to support test automation for mission-based robots,
GzScenic can synthesize mission items (e.g., waypoints, action
locations, the initial position of a robot) as part of a test scenario.
These mission items can be combined with a generated scene via a
developer-provided test harness to allow automated end-to-end
testing (e.g., by spawning the robot at a given initial location,
sending it a generated set of waypoints, and monitoring its progress).

The contributions of this papers are as follows:
\begin{itemize}
\item We introduce GzScenic; a tool that allows users of the popular
  Gazebo simulator to describe
  test scenarios in a the Scenic high-level DSL and automatically generate
  the scenes and missions for the test.
\item We provide an example of using GzScenic for the Fetch robot~\cite{fetch}.
\item We publicly release GzScenic's source code and the example
scenarios at \artifacturl.
\end{itemize}

\section{Background}
\label{sec:background}

\begin{figure*}[!h]
\centering
  \includegraphics[width=0.9\textwidth]{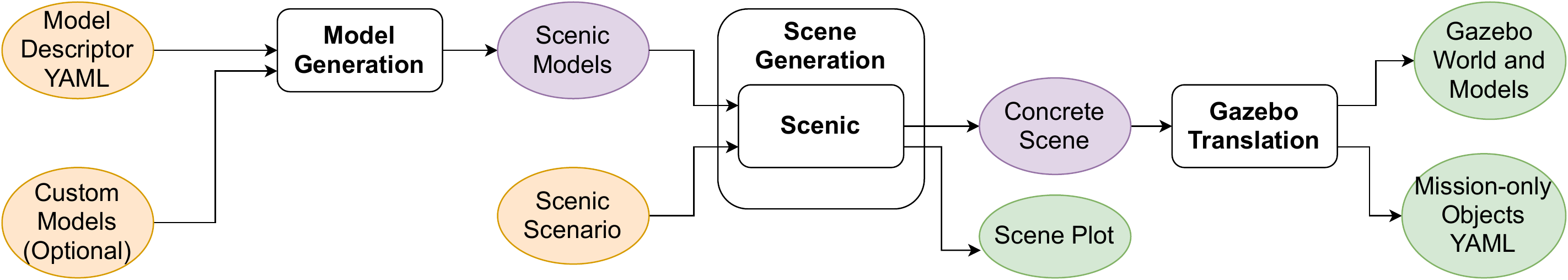}
  \caption{
    \small An overview of GzScenic internal process. Orange ovals represent input provided by the user, purple ovals are internal products, and green ovals are the produced outputs. Rectangles represent the three steps taken by GzScenic.}
  \label{fig:overview}
\end{figure*}

\subsection{Scenic}
\label{sec:scenic}

In this section, we provide a high-level overview of the structure and important 
features of Scenic~\cite{scenic}. We refer the reader to the original Scenic paper
for further details~\cite{scenic}.
Scenic is a domain-specific probabilistic programming language for
modeling the environments of robotic and cyberphysical systems such as
autonomous cars. A Scenic program (i.e., scenario)
defines a distribution over scenes, configurations of physical objects
and agents; sampling from this distribution yields concrete scenes which
can be simulated by the supported simulators. \Cref{fig:scenic}
presents an example Scenic scenario and a concrete scene produced from
that scenario using the GTA V engine.

Overall, Scenic accepts a pre-defined set of models\footnote{Referred
to as \texttt{Classes} in Scenic's documentation.} that define everything
specific to a particular simulator and SUT. For example, in \Cref{fig:scenic},
two instances of the \texttt{Car} model are created.
A portion of the pre-defined \texttt{Car} model is as follows:
\begin{lstlisting}[language=scenic]
class Car:
    position: Point on road
    heading:  roadDirection at self.position
    viewAngle: 80 deg
\end{lstlisting}
which specifies that the position of a \texttt{Car} is a point on
a region called \texttt{road} that is defined separately and represents the 
roads in the GTA map. The car's heading is the same as the
\texttt{roadDirection} that is the nominal traffic direction at 
a point on the road, and its \texttt{viewAngle} is 80 degrees.

To allow Scenic to parse scenarios and generate concrete scenes, a model
must be defined for each entity that can be represented within a given scene.
A concrete scene consists of a set of instantiated models, known as
objects, with concrete values as their properties.
Scenic automatically determines the spatial relationships between
objects in the scene such that they conform to the
specifications of the scenario and do not collide with each
other.\footnote{Users may override this behavior to allow collisions between objects.}
Scenic arranges objects in the scene by treating each objects as a bounding rectangle
on a two-dimensional plane.
At the time of writing, Scenic is unable to arrange objects in three dimensions.

In addition to specifying spatial relationships between objects within a scene,
Scenic can model temporal aspects of scenarios.
For
example, in \Cref{fig:scenic},
we can not only specify where the badly parked car is located, but also how
it should behave over time (e.g., \enquote{pulls into the road as the ego car approaches}).
However, modeling the dynamic scenarios requires direct connection
between Scenic and the simulator, and more complex modeling of
active agents and their behaviors. Since defining these connections
and models are system- and domain-specific, we only focus on generating
static scenes for the rest of the paper.

At the time of writing,
Scenic is compatible with
GTA V~\cite{gta}, CARLA~\cite{CARLA},
Webots~\cite{webots}, and LGSVL~\cite{lgsvl} simulators specifically
used in the
autonomous vehicle sector,\footnote{A simple set of pre-defined models
for a Mars rover in Webots are also included in Scenic~\cite{scenic}.}
and does not support Gazebo; the most popular
general-purpose simulator. Pre-defining Scenic models for a
general-purpose simulator that is commonly used in a wide range of
domains is nearly impossible since each domain requires its own
set of models. GzScenic allows the user to automatically generate
these models by providing a high-level description.

\subsection{Gazebo}
\label{sec:gazebo}

Gazebo is a popular, general-purpose robotics simulator\cite{gazebo,IgnitionGazebo},
maintained by Open Robotics,
that has been used in a wide variety of domains and is the de facto simulation platform
used by ROS.

Running a Gazebo simulation requires several components~\cite{gazebo-components}.
First of all, a world description file
should be provided that describes all the elements in a simulation,
including its objects, robots, sensors, and light sources. This file
typically has a \texttt{.world} extension
and uses the XML-based Simulation Description Format (SDFormat)~\cite{sdf}
to describe those elements.

Included within the world file are model instances, given by
\texttt{<model>} elements, which may be defined directly in the world
file, or, more commonly, included separately by external model files via
the \texttt{<include>} tag.
Defining the model files allow the model to be easily reused among many
worlds. Gazebo model files also follow the SDFormat, and define
all of the components related to modeling an entity such
as joints, collisions, visuals, and plugins.

Included within the components of a model are its collision geometries,
given by \texttt{<collision>} tags, which are used by Gazebo for collision
checking.
These geometries can take on simple shapes such as a box, cylinder, or sphere, or they
can include more complex shapes specified by 3D mesh files, which
can take one of the three supported formats of STL, Collada or OBJ,
with Collada and OBJ being the preferred formats.

\section{GzScenic}
\label{sec:gzscenic}

The goal of GzScenic is to convert a test scenario,
written in Scenic language, to a set of files and models that can be
used by Gazebo. \Cref{fig:overview}
provides an overview of GzScenic's inputs, outputs, and internal
steps.
GzScenic takes a high-level model descriptor YAML file, a set of
custom models, and a Scenic scenario as inputs, and performs
three steps to achieve its goal.
Firstly, it automatically generates Scenic models from the model
descriptor YAML and the custom models (\Cref{sec:model-generation}).
It then passes the generated Scenic models and the input scenario
to the Scenic tool, and generates a concrete scene (\Cref{sec:scene-generation}).
Finally, it translates the generated scene into a format that is
suitable for Gazebo (\Cref{sec:gazebo-translation}).

We provide a running example of generating a scene for the popular
open-source Fetch robot~\cite{fetch}. More examples of GzScenic
inputs and scenarios can
be found in the tool's repository at \artifacturl.
In this example, our goal is to create a scene for Fetch that resembles
a pick and place playground.\footnote{Pick and place is an act
of picking an object, moving it to another location, and placing it at
the destination.}
Throughout the rest of this section, we explain how GzScenic achieves this goal.

\subsection{Model Generation}
\label{sec:model-generation}

\begin{figure}[h]
\centering
  \begin{lstlisting}[language=yaml]
models:
- name: fetch
  type: MISSION_ONLY
  width: 0.57
  length: 0.53
  heading: -1.57
- name: waypoint
  type: MISSION_ONLY
- name: cafe_table
  type: GAZEBO_MODEL
- name: bookshelf
  type: GAZEBO_MODEL
- name: LampAndStand
  type: GAZEBO_MODEL
- name: demo_cube
  type: CUSTOM_MODEL
  dynamic_size: False
models_dir: models/
world: empty_world.world
\end{lstlisting}
  \caption{
    \small An example model descriptor YAML file for
    Fetch.}
  \label{fig:yml-example}
\end{figure}

As discussed in \Cref{sec:scenic}, parsing a scenario
in Scenic language
and generating a valid concrete scene
requires a set of model definitions that should be provided to Scenic.
These models should describe all entities that can be included in a scenario.
For example, in self-driving applications these models may include
entities such as cars, roads, and pedestrians. Scenic provides
some of the models out of the box for self-driving applications,
which are its primary domain.

Since Gazebo is a general-purpose
simulator that is used in a wide range of domains,
it is nearly impossible to pre-define Scenic models that describe
the entities required for simulation of all systems in different sectors. For example, an agricultural robot
requires modeling of entities such as plants and tractors,
whereas a warehouse robot requires modeling of the shelves, boxes,
and rooms.
In comparison, defining these models for a domain-specific simulator such as
GTA V and CARLA requires a one-time investment since most of the
entities that can be simulated and included in the scenarios are
shared among all systems that use these simulators.
For example,
if models are produced for CARLA in order to test a given system,
those same models may be reused in another system with minimal effort.

GzScenic allows Gazebo users to easily create Scenic models by automatically
generating them from a set of Gazebo models, provided as \texttt{.sdf}
and 3D mesh (e.g., \texttt{.dae}, \texttt{.obj}, \texttt{.stl}) files,
as described in \Cref{sec:gazebo}.
To perform this conversion, GzScenic requires that the user to provide a
list of the models that may be used in generated scenes via the YAML
model descriptor file, illustrated in \Cref{fig:yml-example}.
The description of each model in the file should specify its name,
and its type.
The three model types, described below, inform GzScenic of how it should access the
Gazebo models (if required).

\begin{itemize}
\item \texttt{GAZEBO\_MODEL}:
By default, Gazebo comes prepackaged with a common database of
models.\footnote{\url{https://github.com/osrf/gazebo_models}}
In addition to this database, Ignition Fuel web application hosts
thousands of Gazebo models publicly released by users.\footnote{\url{https://app.ignitionrobotics.org/fuel/models}}
Models of type \texttt{GAZEBO\_MODEL} refer to these models.
GzScenic automatically downloads
all the files related to models of this type from the model 
distribution according to the provided name.
\item \texttt{CUSTOM\_MODEL}: Models of this type are not standard Gazebo 
models. They are either made by the user for their own use, or should be downloaded from a custom source.
In the former case,
GzScenic looks for the Gazebo model files in the \texttt{models\_dir} directory
specified by the model descriptor file, and in the later case, GzScenic downloads
the files from the URL provided by the tag \texttt{url} in the YAML file.
\item \texttt{MISSION\_ONLY}: Models of this type include any entities
in the scenarios that
do not map to a simulated object that should be included in the
Gazebo \texttt{.world} file, but
are particularly important in generating interesting missions in the
scenarios. For example, mission waypoints are entities
that do not represent objects in the environment, but may be
specified in a scenario to allow missions to be generated and executed by a
test harness.
Another example is the robot itself. Robots are not typically included in the
\texttt{.world} file, and are spawned separately by \texttt{roslaunch}.
Therefore, these robots should not be included in the generated \texttt{.world}
file, but a suitable initial position should be emitted by GzScenic to allow
users to test the robot in different, valid starting positions.
We discuss the use of GzScenic to generate missions further in \Cref{sec:gazebo-translation}.
\end{itemize}

\Cref{fig:yml-example} presents an example model descriptor file for
Fetch, describing the models that may be used
in a pick and place scenario.
In this list, we have included models for \texttt{cafe\_table}, \texttt{bookshelf},
provided out of the box by the official Gazebo model database,
and \texttt{LampAndStand}, released on Ignition Fuel web application.
The Gazebo model
for \texttt{demo\_cube} is custom-made and provided in the \texttt{models/} directory.
Finally, we include model descriptions for the robot and waypoints,
\texttt{fetch} and \texttt{waypoint}, as \texttt{MISSION\_ONLY}.
Note that this is only
an example of the set of models that can be included in the scenarios.
Users can select their preferred models from thousands
of available models, or including their own custom-made models.

To generate corresponding Scenic models for the models in the model descriptor file,
GzScenic automatically determines a number of features for each model.
These features include a 2D bounding box,\footnote{Recall that Scenic treats all models as 2D rectangles.}
given by a \texttt{length} and \texttt{width},
and a flag, \texttt{dynamic\_size}, indicating whether or not the model can be dynamically
resized.
To determine the \texttt{length} and \texttt{width} of a model,
GzScenic computes a bounding box for each individual collision geometry specified
in the \texttt{.sdf} file, before determining a bounding box for the entire model.
Note that GzScenic supports 5 of the 9 types of collision geometry that
can be represented using SDF~\cite{sdf}: \texttt{empty}, \texttt{box},
\texttt{cylinder}, \texttt{sphere}, and \texttt{mesh}. Of the 283
models that are prepackaged with Gazebo, only 12 use a
geometry that is not supported by GzScenic.

After calculating the bounding box of the model, GzScenic
determines whether the model can be resized.
For example, a simple box, a tree, or a wall should be allowed
to be resized based on what the scenario requires, but a table that
consists of multiple parts (e.g., surface and legs), a robot model,
or a stop light should not be resized since they are not scalable in
the real world. As a rule of thumb, GzScenic allows dynamic resizing
down to half or up to twice the original size of the models
if they only consist of a single simple collision geometry (i.e.,
\texttt{empty}, \texttt{box}, \texttt{cylinder}, or \texttt{sphere}).
We made this decision
based on the observation that complex models that include multiple
collision geometries or meshes are more likely to be of standard size and
should not be resized. However, there are exceptions to this rule.
As a result, GzScenic allows the user to override \texttt{dynamic\_size}
feature of a model in the model descriptor file.

Once all of the models specified in the model descriptor file are transformed
to Scenic models, Scenic can interpret the input scenario.
Note that this model generation step need only occur once
for each system unless the models change.
GzScenic stores generated models for future use.

\subsection{Scene Generation}
\label{sec:scene-generation}

In this step, GzScenic runs the Scenic interpreter on the provided scenario description
using the models generated in the previous step.
Scenic, if possible, generates a concrete scene
that satisfies the scenario description, and shows a
plot of object arrangements on a 2D plane to the user (\Cref{fig:example-plot}).

\begin{figure}
\centering
\small
\begin{lstlisting}[language=scenic]
width = 8
length = 8
heading = 0
workspace = Workspace(RectangularRegion(0 @ 0, heading, width, length))

create_room(length, width, x=0, y=0, sides='NSWE')

ego = Fetch at 0 @ 0

table1 = CafeTable offset by 0 @ 1, facing 0 deg
create_room(3, 2.5, x=-2, y=2, sides='NSE')
table2 = CafeTable at -2 @ 2
Bookshelf at Range(-4,4) @ -3.5, facing 180 deg
back_right_region = RectangularRegion(-2 @ -2, 0, 3.5, 3.5)
Lampandstand in back_right_region
\end{lstlisting}
\caption{An example scenario for Fetch, written in Scenic.}
\label{fig:example-scenario}
\end{figure}


\Cref{fig:example-scenario} presents a simple example
scenario for the Fetch robot. In this scenario, we first
create a workspace\footnote{A workspace in Scenic language specifies
the region the objects must lie within.} with length and width of 8 meters.
Then, using a function provided by GzScenic, \textbf{create\_room},
we create walls surrounding the workspace on all four sides.
The rest of the scenario includes description of the objects and
their positioning in the scene, and creation of a smaller room that
has walls on three sides. The resulting scene plot generated by Scenic
is presented in \Cref{fig:example-plot}.
The instance in the center
of the plot is the ego, which is the Fetch robot in this scenario.
All other instances are represented as red rectangles. Note that the
position and orientation of 3 of the instances in this scenario are
randomly determined and can take other concrete values in other
scenes.

The output of this step is a concrete scene,
that includes all the
objects generated from the models, and their arrangements.
Next, we translate this concrete scene into a format that can be
used by Gazebo.

\subsection{Gazebo Translation}
\label{sec:gazebo-translation}

This final step of GzScenic accepts a concrete scene as an input,
and translates this scene into a) Gazebo world and models that
allow us to simulate the scene in Gazebo, and b) a YAML file
listing the position and orientation of \texttt{MISSION\_ONLY} objects,
which can facilitate automated creation of test missions.

\begin{figure*}
\centering
    \begin{subfigure}{0.72\columnwidth}
    \includegraphics[width=\columnwidth]{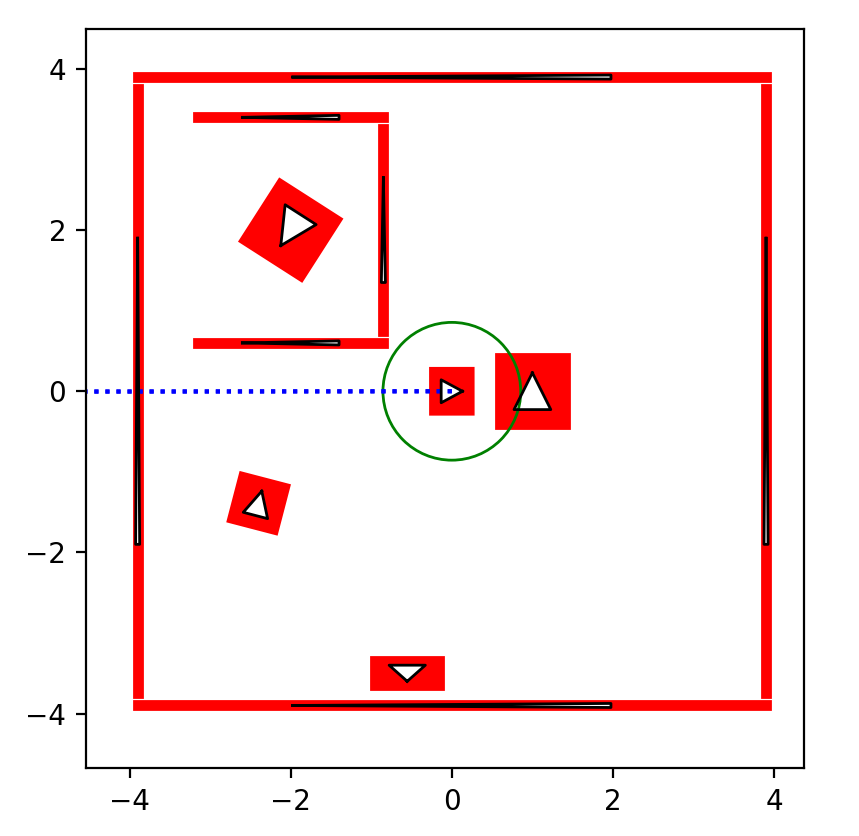}
    \caption{\small 2D plot generated for the concrete scene.}
    \label{fig:example-plot}
    \end{subfigure}
    \hspace{10mm}
    \begin{subfigure}{\columnwidth}
    \centering
	\includegraphics[width=1.1\columnwidth]{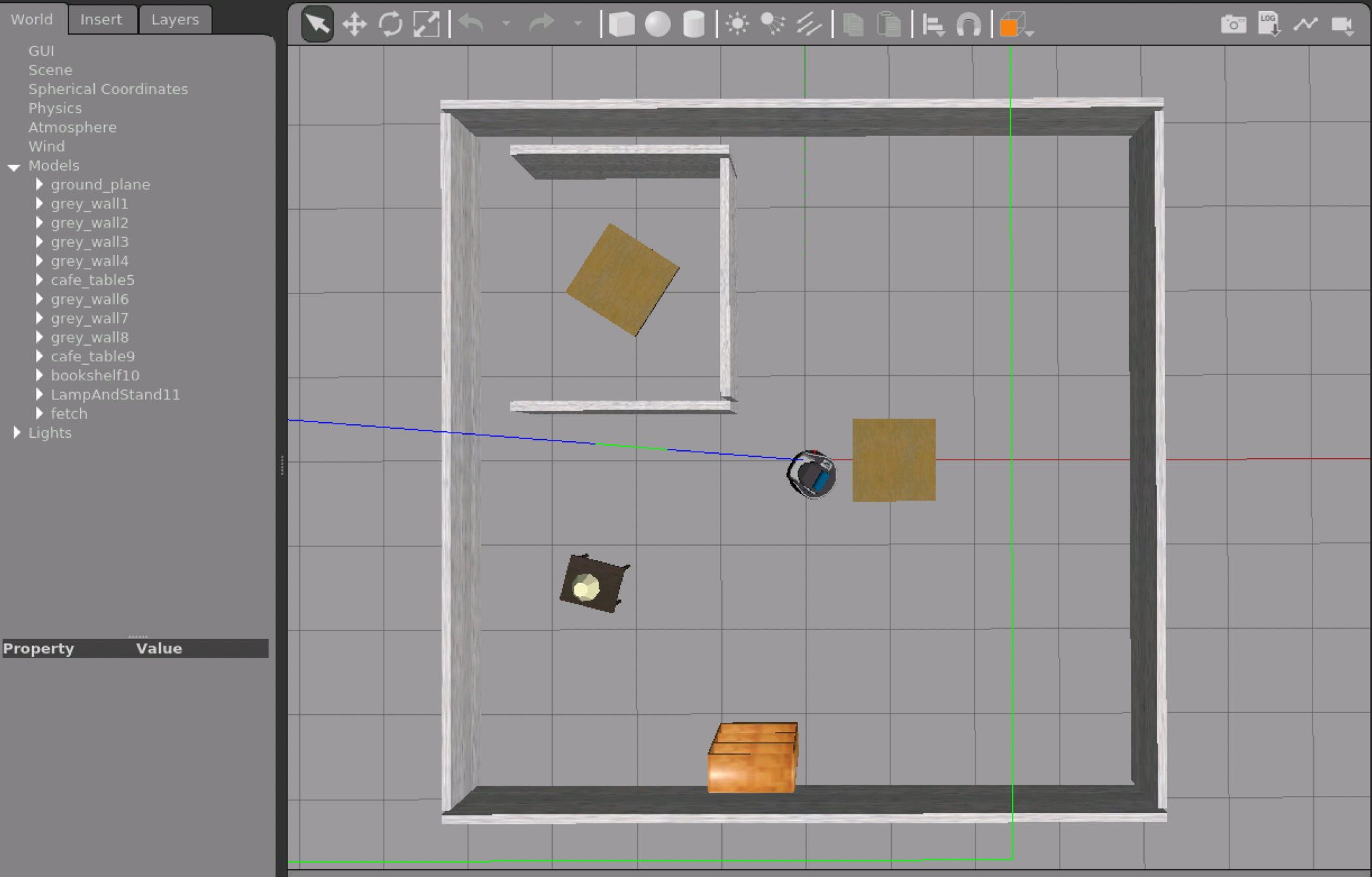}
    \caption{\small Gazebo simulation of the generated scene.}
    \label{fig:example-gazebo}
    \end{subfigure}
    \caption{\small An example of the scene generated for the
    Fetch robot reflecting the scenario
    of \Cref{fig:example-scenario}.}
    \label{fig:example}
\end{figure*}

\paragraph{Gazebo world and models}
As mentioned in \Cref{sec:gazebo}, Gazebo components
include
a world file (usually with \texttt{.world} extension) and a set of
models in the form of \texttt{.sdf} files, mesh files, and configuration files~\cite{gazebo-components}.
In this step, GzScenic translates a concrete scene description
to corresponding Gazebo \texttt{.world} and model files.
To do so, GzScenic starts from an empty world environment, which,
by default, includes only a ground plane, and adds every object
in the concrete scene to this world one by one. 
The user can provide a customized empty world to GzScenic where
they can configure different aspects of the world such as its lighting, shadowing, and physics engine.

For every object in a concrete scene, GzScenic determines the
elements that must be added to the world file, and the files that
need to accompany the generated world.
For all objects generated from a \texttt{GAZEBO\_MODEL} or \texttt{CUSTOM\_MODEL},
GzScenic adds those objects to the world file via the \texttt{<include>} tag.
Additionally,
GzScenic generates the necessary Gazebo model files for
each individual object where the collision geometries must be
updated to reflect the dynamically-determined size of the object.

At the end of the process, the user ends up with a world file and a set of models.
If GzScenic is running on the same system as Gazebo,
the user can specify the output directory in such a way that
Gazebo can immediately find the 
GzScenic's outputs.
Note that the path to the models directory can be passed
to Gazebo via the \texttt{GAZEBO\_MODEL\_PATH} environment variable.
If GzScenic is not running on the same system as Gazebo,
the user must transfer the output of GzScenic to the system that hosts
Gazebo.

In our example scenario (\Cref{fig:example-scenario}),
GzScenic automatically translates
the concrete scene plotted in \Cref{fig:example-plot}
to a Gazebo simulation presented in 
\Cref{fig:example-gazebo}.
As shown, the position and orientation of the objects
in this simulation are aligned with the generated plot, and the
description provided in the scenario of \Cref{fig:example-scenario}.

\paragraph{Mission-only objects YAML}

Let us refer back to the example
scenario of \Cref{fig:example-scenario} for the Fetch robot.
Our ultimate goal in this case is to test
Fetch in a scene that is generated from our example scenario.
Simply launching Fetch in the automatically-generated Gazebo simulation
(\Cref{fig:example-gazebo}) will not test the system as it is
not performing any operations. The system should receive a mission
(i.e., a set of instructions to perform actions) to be tested in this
environment. For example, a mission for Fetch can instruct the
robot to pick an object from the table, move to another room, and place
the object on the other table.

A scenario may include information about the mission that should be
performed in the generated scene. For example, in the scenario of
\Cref{fig:example-scenario}, we can add instances of
\texttt{waypoint}s that reflect where the robot should move to:
\begin{lstlisting}[language=scenic]
Waypoint in back_right_region
Waypoint ahead of table2 by 1
\end{lstlisting}

GzScenic automatically generates position and orientation for these
instances. However, since their model type is \texttt{MISSION\_ONLY}
it does not include these objects in the Gazebo simulation. Instead,
it outputs a YAML file that lists the position and
orientation of each one of these \texttt{MISSION\_ONLY} objects,
grouped by their type:
\begin{lstlisting}[language=yaml]
fetch:
- heading: -1.57
  x: 0
  y: 0
  z: 0.0
waypoint:
- heading: 4.521433387130848
  x: -0.6426244725245782
  y: -0.7777737656890915
  z: 0.0
- heading: 2.2663887353720784
  x: -2.7676780355847423
  y: 1.3591564670641116
  z: 0.0
\end{lstlisting}

Since the definition of the missions and how they are executed are
system-specific, there is no generic way to convert this list of
coordinates to a running mission that will work on all systems.
However, we believe that users can easily read this
output file to automatically generate their intended missions using a
custom test harness.

%

\section{Limitations}

As mentioned in \Cref{sec:scenic}, Scenic is only capable of
generating 2D scenes and cannot arrange objects in the scene in
a 3D environment. This creates a limitation for GzScenic as well
for scenarios that for example require multiple objects stacked on
top of each other.

While resolving this limitation is out of the scope
for GzScenic, there is a workaround that will allow users to
stack objects on top of each other in GzScenic. GzScenic by default 
keeps track of the height and z coordinate of the objects. This 
information has no impact on the scene that is generated by Scenic but
is used during the Gazebo translation step to create Gazebo models.
To stack two objects on top of each other, the user need to properly
set the \texttt{z} value of the objects, and allow them to collide by
setting \texttt{allowCollisions} to \texttt{True}. In the example
scenario of Figure~\ref{fig:example-scenario}, we can place a cube on the
table by adding the following lines to the scenario:
\begin{lstlisting}[language=scenic]
table.allowCollisions = True
cube = Cube at table.position, with allowCollisions (True)
cube.z = table.height + cube.height
\end{lstlisting}
Note that this is only a temporary workaround until handling 3D
positions is added to Scenic.

Another limitation of GzScenic that arises from Scenic features is the
fact that all objects are considered as rectangles in the 2D space.
As a result GzScenic computes the bounding box of models as described
in \Cref{sec:model-generation}. However, the bounding box of a
model is not always truly representative of the space that the model
is going to take. For example, imagine a hoop that is hollow inside.
The bounding box of this hoop will be considered as a square surrounding
its circumference, and the hoop is treated the same as a solid box by
Scenic. However, we may want to allow an object to be placed in the
center of the hoop, which is currently not allowed.
To partially mitigate this issue, 
we plan to improve GzScenic to break large models
(e.g., model of a house) into smaller ones instead of creating a 
bounding box for the whole model.

\section{Conclusion}

In this work we present GzScenic, a tool that automatically generates
scenes for the Gazebo simulation from scenarios provided in the
Scenic language. GzScenic allows the users to simply specify a list
of models they intend to use in the simulation, and it automatically
turns these models into models that are interpretable by Scenic. Using 
these models, Scenic generates a scene from the scenario, which later
on is automatically converted to Gazebo models by GzScenic.

\section*{Acknowledgement}

This research was partly funded by AFRL (\#OSR-4066).
The authors are grateful for their support.
Any opinions, or findings expressed are those of the authors
and do not necessarily reflect those of the US Government.


\bibliographystyle{IEEEtran}
\bibliography{iros}

\end{document}